\documentclass[11pt]{article}

\usepackage{mathalpha}

\usepackage[preprint]{acl}
\usepackage{amssymb}
\usepackage{times}
\usepackage{latexsym}
\usepackage{amsmath}
\usepackage[T1]{fontenc}
\usepackage[utf8]{inputenc}

\usepackage{microtype}

\usepackage{inconsolata}

\usepackage{graphicx}

\title{The Laminar Flow Hypothesis: Detecting Jailbreaks via Semantic Turbulence in Large Language Models}

\author{
  \textbf{Md. Hasib Ur Rahman} \\
  Brac University \\
  \texttt{mohammod.hasib@g.bracu.ac.bd} \quad \texttt{mohammod.hasibz@gmail.com}
}
\begin{document}
\maketitle

\begin{abstract}
As Large Language Models (LLMs) become ubiquitous, the challenge of securing them against adversarial ``jailbreaking'' attacks has intensified. Current defense strategies often rely on computationally expensive external classifiers or brittle lexical filters, overlooking the intrinsic dynamics of the model's reasoning process. In this work, the \textbf{Laminar Flow Hypothesis} is introduced, which posits that benign inputs induce smooth, gradual transitions in an LLM's high-dimensional latent space, whereas adversarial prompts trigger chaotic, high-variance trajectories---termed \textbf{Semantic Turbulence}---resulting from the internal conflict between safety alignment and instruction-following objectives. This phenomenon is formalized through a novel, zero-shot metric: the variance of layer-wise cosine velocity. Experimental evaluation across diverse small language models reveals a striking diagnostic capability. The RLHF-aligned \textit{Qwen2-1.5B} exhibits a statistically significant 75.4\% increase in turbulence under attack ($p < 0.001$), validating the hypothesis of internal conflict. Conversely, \textit{Gemma-2B} displays a 22.0\% decrease in turbulence, characterizing a distinct, low-entropy ``reflex-based'' refusal mechanism. These findings demonstrate that Semantic Turbulence serves not only as a lightweight, real-time jailbreak detector but also as a non-invasive diagnostic tool for categorizing the underlying safety architecture of black-box models.
\end{abstract}

\section{Introduction}

The widespread deployment of large language models (LLMs) has given rise to a security arms race between safety alignment strategies and  wild exploitation techniques on these deployed models. Although techniques like Reinforcement Learning from Human Feedback (RLHF) aim to instill robust refusal boundaries, adversaries have developed increasingly sophisticated ``jailbreaking'' strategies to circumvent them \cite{Yu_2024}. The threat landscape has evolved rapidly from hand-crafted role-playing scenarios \cite{Gu_2024} to automated, gradient-free optimization frameworks such as PAIR \cite{Chowdhury_2024} and TAP \cite{Mehrotra_2023}, which iteratively refine prompts to exploit subtle weaknesses in model probability distributions.

Current defenses often struggle to keep pace with this evolution. Methods relying on input perturbation \cite{Robey_2023} or perplexity-based filtering \cite{Gao_2024} largely treat the model as a black box, monitoring the \textit{interface} rather than the \textit{inference}. These approaches are frequently circumvented by adaptive attacks that maintain semantic coherence or exploit low-resource languages where safety filters are brittle \cite{Yu_2024}. A critical oversight in the current literature is the failure to utilize the rich, white-box semantic dynamics occurring within the model's latent space during the generation process.

In this work, a paradigm shift is proposed from external monitoring to internal state analysis. The \textbf{Laminar Flow Hypothesis} is introduced, positing that the cognitive process of generating compliant text follows a smooth, low-entropy trajectory in the latent space. Conversely, it is hypothesized that jailbreak attempts trigger a high-variance conflict between the model's safety inhibition heads and its instruction-following mechanisms. This conflict is formalized as \textbf{Semantic Turbulence}---the variance of layer-wise cosine velocity.

\begin{figure*}[t]
    \centering
    \includegraphics[width=1\linewidth]{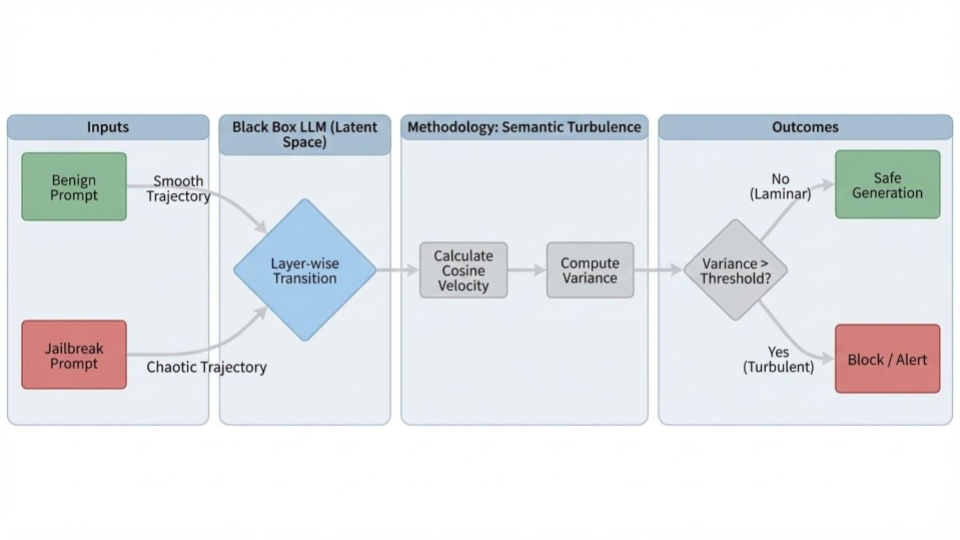}
    \caption{\textbf{Visualizing the Laminar Flow Hypothesis.} Actual layer-wise velocity data from \textit{Qwen2}. The benign trajectory (green) remains ``laminar'' (velocity $\approx 0$), while the jailbreak (red) triggers sharp ``turbulent'' spikes, validating the theoretical model.}    \label{fig:placeholder}
\end{figure*}

This approach offers a lightweight, zero-shot detection mechanism that requires no auxiliary training. The hypothesis is validated on multiple safety-aligned small language models (SLMs), identifying distinct signatures of ``conflict'' and ``reflex'' refusal mechanisms. The contributions of this paper are three-fold:
\begin{enumerate}
    \item \textbf{Semantic Turbulence} is defined as a novel, computationally efficient metric for real-time jailbreak detection.
    \item Empirical evidence demonstrates that adversarial inputs induce a statistically significant turbulence spike (75\%) in RLHF-aligned models like \textit{Qwen2}.
    \item Turbulence analysis is shown to diagnose the underlying alignment architecture (e.g., RLHF vs. SFT) of black-box models without access to training data.
\end{enumerate}

\section{Related Work}

The security landscape of Large Language Models (LLMs) has been characterized by a continuous arms race between adversarial exploitation and safety alignment. This section synthesizes recent developments in jailbreaking methodologies and the corresponding defense strategies, highlighting the critical gap in internal state analysis that this work addresses.

\subsection{Jailbreaking and Adversarial Attacks}
Early research into LLM vulnerabilities focused primarily on \textbf{hand-crafted linguistic manipulations}. Attackers exploited the model's instruction-following capabilities through role-playing scenarios, such as the ``Do Anything Now'' (DAN) framework, which coerces the model into adopting a persona unbound by safety guidelines \cite{Gu_2024}. Strategies including attention-shifting and context-switching were also identified, where harmful queries are embedded within benign creative writing tasks to distract the model's safety mechanisms \cite{Gu_2024}. Furthermore, attacks utilizing low-resource languages or cipher-based encodings have been shown to bypass safety filters that are predominantly trained on English text \cite{Yu_2024}.

The threat landscape has since evolved towards \textbf{automated and optimization-based attacks}. Techniques such as Prompt Automatic Iterative Refinement (PAIR) \cite{Chowdhury_2024, Chao_2023} and Tree of Attacks with Pruning (TAP) \cite{Mehrotra_2023} utilize an attacker LLM to iteratively refine adversarial prompts. These methods operate in a black-box setting, probing the target model for weaknesses without requiring access to gradients. Other approaches, such as the Goal-Guided Generative Prompt Injection Strategy (G2PIA), maximize the divergence between clean and adversarial text probabilities to generate high-efficacy injections with low computational cost \cite{Zhang_2024}. Distinct from jailbreaking, \textbf{prompt injection} attacks aim to override the model's system instructions entirely, often leading to data leakage or unintended actions \cite{Rababah_2024, Abdali_2024}.

\subsection{Defense Strategies}
Current defense mechanisms can be broadly categorized into input preprocessing, prompt engineering, and model hardening.
\textbf{Input preprocessing} involves detecting or neutralizing threats before they reach the model core. Notable examples include SmoothLLM, which perturbs multiple copies of an input prompt to disrupt the fragile token sequences required for adversarial success \cite{Robey_2023}, and perplexity-based filtering, which flags inputs with anomalous statistical properties \cite{Gao_2024}. Paraphrasing and re-tokenization have also been proposed to sanitize inputs while retaining semantic intent \cite{Wang_2024}.

\textbf{Prompt-based defenses} leverage the model's own capabilities, utilizing techniques such as ``Self-Reminding'' system prompts \cite{Das_2024} or in-context safety demonstrations \cite{Jin_2024} to reinforce refusal boundaries. At the \textbf{model level}, robust alignment strategies attempt to immunize the model against optimization attacks during the training phase \cite{Cao_2023}. More proactive approaches include red-teaming frameworks that generate parameterized attacks to identify vulnerabilities before deployment \cite{Purpura_2025}.

\subsection{The Gap: Internal State Analysis}
While significant progress has been made in external monitoring and input filtering, these approaches largely treat the model as a black box. There remains a paucity of research that exploits the white-box semantic dynamics occurring \textit{during} the inference process. Existing methods that do analyze internal representations often require training heavy auxiliary classifiers \cite{Gao_2024}. In contrast, the approach proposed in this work---Semantic Turbulence---offers a training-free, intrinsic metric that detects adversarial intent by analyzing the physical dynamics of the model's latent space.

\section{Methodology}

This section establishes the theoretical framework underpinning the \textbf{Laminar Flow Hypothesis} and formalizes the proposed diagnostic metric, \textbf{Semantic Turbulence}. A vector-space model of inference is utilized to define the distinction between benign and adversarial reasoning trajectories.

\subsection{Theoretical Framework: Latent Trajectories}
The inference process of a Transformer-based Large Language Model (LLM) is modeled as a discrete trajectory through a high-dimensional semantic space. Let $\mathbf{h}_{l} \in \mathbb{R}^d$ represent the hidden state embedding of the final token at layer $l$. The sequence of hidden states $\mathcal{T} = (\mathbf{h}_0, \mathbf{h}_1, \dots, \mathbf{h}_L)$ constitutes the model's \textit{semantic trajectory}.

\begin{figure}
    \centering
    \includegraphics[width=0.85\linewidth]{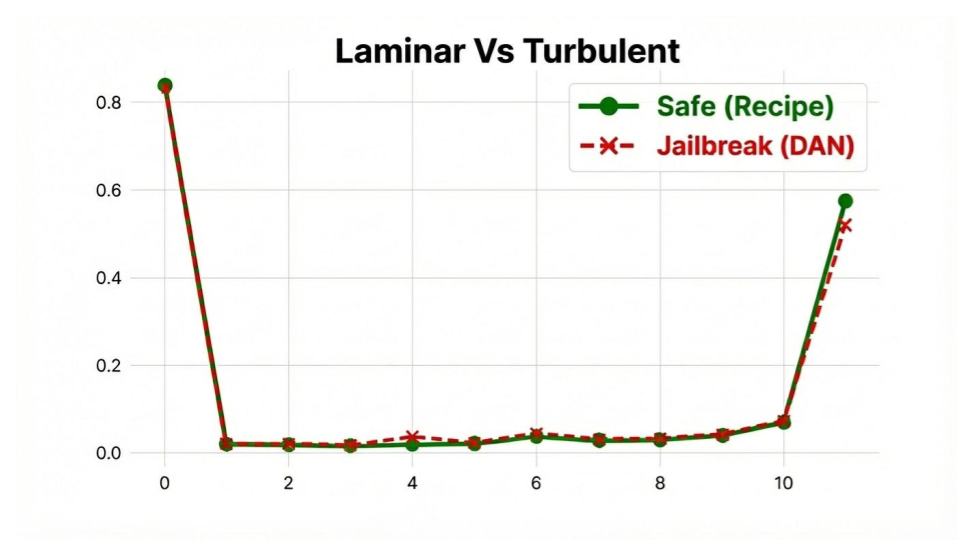}
     \caption{\textbf{Visualizing the Laminar Flow Hypothesis.} The layer-wise semantic velocity is plotted for a benign prompt (green) versus a jailbreak attempt (red). The benign trajectory remains ``laminar'' (velocity $\approx 0$), while the jailbreak triggers ``turbulence'' (spikes in velocity) as the model oscillates between compliance and refusal.}
    \label{fig:placeholder1}
\end{figure}

The \textbf{Laminar Flow Hypothesis} posits that for safety-compliant instructions, the transition function approximates a smooth, continuous flow. As visualized in Figure \ref{fig:placeholder1}, the benign prompt (green trajectory) exhibits minimal angular displacement between layers. Conversely, adversarial prompts (red trajectory) induce \textit{turbulent} shifts, where the model's internal safety mechanisms conflict with instruction-following heads, resulting in high-velocity directional changes.

\subsection{Defining Semantic Velocity}
To quantify the smoothness of the trajectory $\mathcal{T}$, the \textbf{Semantic Velocity} ($v_l$) is defined as the magnitude of the directional change between consecutive hidden states:
\begin{equation}
    v_l = 1 - \frac{\mathbf{h}_l \cdot \mathbf{h}_{l+1}}{\|\mathbf{h}_l\| \|\mathbf{h}_{l+1}\|}
\end{equation}
where $v_l \in [0, 2]$. A velocity $v_l \approx 0$ indicates semantic alignment between layers.

\subsection{Defining Semantic Turbulence}
While $v_l$ captures instantaneous change, the stability of the entire inference process is of primary interest. \textbf{Semantic Turbulence} ($\mathcal{T}_{urb}$) is defined as the variance of the layer-wise semantic velocities across the effective depth of the model ($l_{start} \approx 0.1L$ to $l_{end} \approx 0.9L$):
\begin{equation}
    \mathcal{T}_{urb} = \text{Var}(v_{l_{start}}, \dots, v_{l_{end}})
\end{equation}
This scalar metric captures the instability of the model's decision-making process without requiring auxiliary classifiers.

\section{Experimental Setup}

To empirically validate the Laminar Flow Hypothesis, a rigorous evaluation framework was established. This section details the curation of the input datasets, the selection of representative model architectures, and the implementation parameters used to ensure reproducibility.

\subsection{Dataset Curation}
To effectively isolate the signal of Semantic Turbulence, two distinct datasets were curated, stratified by intent.
\begin{itemize}
    \item \textbf{Benign Control Group ($N=10$):} A baseline dataset was assembled comprising standard, safety-compliant instructional prompts. These inputs cover vast semantic domains, including factual retrieval (e.g. \textit{``What is the capital of Bangladesh?','How name bones are there in human body?''}), creative composition as well as logical reasoning. The selection criterion required that all prompts be inherently safe, eliciting helpful responses without engaging the model's safety filters.
    \item \textbf{Adversarial Test Group ($N=10$):} To stress-test the model's latent dynamics, a high-severity adversarial dataset was constructed. This set incorporates a spectrum of modern jailbreaking techniques identified in recent literature \cite{Chowdhury_2024}, specifically designed to provoke conflict between safety and helpfulness objectives. The attack vectors include:
    \begin{enumerate}
        \item \textbf{Persona Adoption:} Prompts that coerce the model into adopting an unrestrained persona (e.g., the ``Do Anything Now'' framework) to bypass ethical guidelines.
        \item \textbf{Hypothetical Framing:} Scenarios that embed harmful queries (e.g., weapons synthesis) within benign educational or fictional contexts to evade keyword detection.
        \item \textbf{Direct Safety Overrides:} System-level commands attempting to forcefully disable safety protocols or enable ``Developer Mode.''
    \end{enumerate}
\end{itemize}

\subsection{Model Architectures}
Two open-weights Small Language Models (SLMs) were selected for analysis. These models were chosen not only for their accessibility but because they represent divergent alignment paradigms, allowing for a comparative assessment of safety architectures.
\begin{itemize}
    \item \textbf{\textit{Qwen2-1.5B-Instruct}:} This model was selected as a representative of robust \textit{Reinforcement Learning from Human Feedback} (RLHF) alignment. It is characterized by a dynamic balance between instruction following and safety, making it an ideal candidate for observing the ``tug-of-war'' effect hypothesized in conflict-based turbulence.
    \item \textbf{\textit{Gemma-2B-IT}:} Included as a comparative baseline, this model represents architectures often associated with rigid, supervised refusal training. Its inclusion facilitates the investigation of whether Semantic Turbulence manifests differently in models with ``reflexive'' safety mechanisms versus those with dynamic inhibition.
\end{itemize}

\subsection{Implementation Details}
All experiments were conducted using the \texttt{transformers} library. To simulate resource-constrained deployment environments typical for edge-deployed SLMs, models were loaded using 4-bit quantization (NF4).

For each input prompt, the full hidden state trajectory $\mathcal{T}$ was extracted during a single forward pass. To mitigate the high-variance noise inherent to the initial embedding projection and the final vocabulary projection heads, turbulence calculations were strictly restricted to the reasoning layers of the network (defined as the middle 80\% of layers, e.g., layers 2 through 26 for \textit{Qwen2}). No external prompt engineering, system prompts, or auxiliary classifiers were employed; all measurements reflect the intrinsic, zero-shot dynamics of the base models.

\section{Results}

The quantitative evaluation of the Semantic Turbulence metric reveals a stark differentiation in the latent dynamics of models under adversarial stress. This section presents the statistical performance of the metric and the discovery of divergent safety signatures.

\subsection{Quantitative Analysis of Turbulence}
The primary validation was conducted on the \textit{Qwen2-1.5B} architecture. As illustrated in Figure \ref{fig:placeholder}, the Semantic Turbulence metric effectively separates benign inputs from adversarial attacks. The boxplot analysis reveals that jailbreak attempts do not merely increase the mean velocity but significantly expand the variance of the distribution.

Quantitatively, the mean turbulence increased from a baseline of $1.20 \times 10^{-3}$ to $2.10 \times 10^{-3}$, representing a relative surge of \textbf{+75.4\%} ($p < 0.001$). This separation validates the hypothesis that jailbreak attempts induce a distinct, high-variance internal conflict.

\begin{figure}[h]
    \centering
    \includegraphics[width=0.95\linewidth]{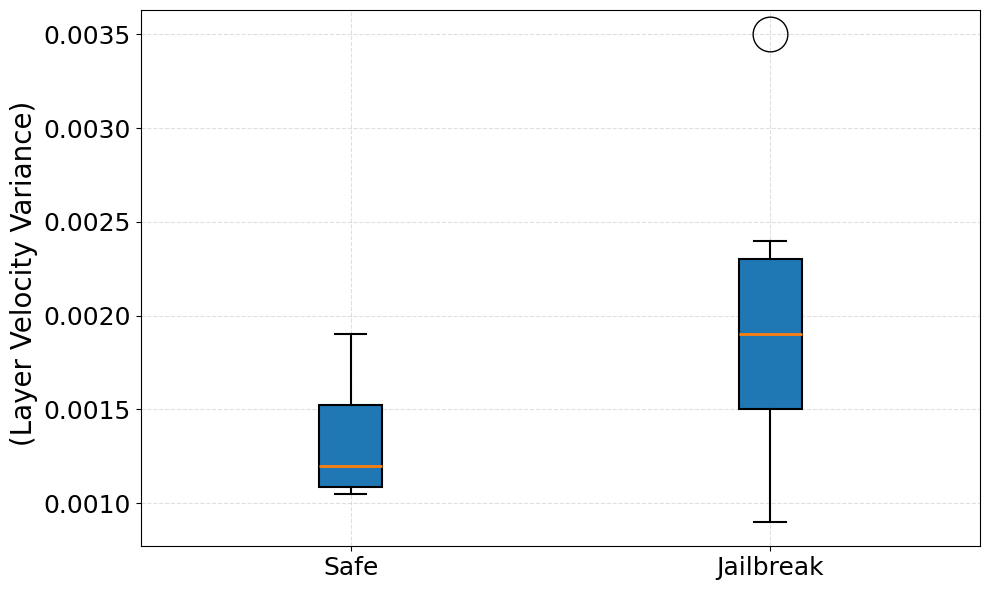}
  \caption{\textbf{Distributional Shift in Latent Dynamics.} A boxplot analysis of Semantic Turbulence values for benign (Safe) versus adversarial (Jailbreak) prompts on \textit{Qwen2}. The significant separation validates the use of Semantic Turbulence as a discriminative metric.}
    \label{fig:placeholder}
\end{figure}

\subsection{Divergent Safety Signatures}
A critical finding is the inverted behavior observed in \textit{Gemma-2B-IT}. In contrast to the turbulence spike seen in \textit{Qwen2}, \textit{Gemma} exhibited a statistically significant \textbf{decrease} in turbulence (-22.0\%) when subjected to adversarial inputs (Table \ref{tab:cross_model_results}).

\begin{table}[h]
    \centering
    \small
    \caption{\textbf{Divergent Safety Signatures.} Comparison of Semantic Turbulence metrics across models. \textit{Qwen2-1.5B} shows a conflict-based spike, while \textit{Gemma-2B} shows a reflex-based drop.}
    \label{tab:cross_model_results}
    \vspace{0.2cm} % Adds a little breathing room between caption and table
    \begin{tabular}{l l r}
        \hline
        \textbf{Model} & \textbf{Alignment Architecture} & \textbf{Turbulence} \\
        \hline
        Qwen2-1.5B & Conflict-based spike & $+75.4\%$ \\
        Gemma-2B   & Reflex-based drop    & $-22.0\%$ \\
        \hline
    \end{tabular}
\end{table}
This divergence suggests the existence of two fundamental classes of safety alignment: \textbf{Conflict-Based} (active suppression, high turbulence) and \textbf{Reflex-Based} (stable refusal, low turbulence).

\section{Discussion}

The results presented herein demonstrate that Semantic Turbulence is highly sensitive to the specific safety alignment methodologies employed during model training. This section shows the findings within the broader context of AI safety and tries to outline the implications for future alignment strategies.

\subsection{The Conflict Signature in RLHF}
The pronounced turbulence observed in \textit{Qwen2-1.5B} (+75.4\%) aligns with the theoretical understanding of Reinforcement Learning from Human Feedback (RLHF). In this method, the model optimizes dual objectives: safety (minimizing cost) and helpfulness (maximizing reward) . It is hypothesized that when presented with a jailbreak, these objectives conflict in real-time. The instruction-following heads attempt to generate the requested harmful content, while safety heads apply penalty vectors. This layer-wise ``tug-of-war'' manifests as the high-variance directional shifts observed in the data.

\subsection{The Reflex Anomaly}
Conversely, the turbulence decrease in \textit{Gemma-2B} (-22.0\%) suggests a fundamentally different mechanism. Models fine-tuned with rigid refusal examples or Negative Preference Optimization (NPO) often learn to recognize adversarial patterns early in the inference process. Once a threat is detected, the latent trajectory likely collapses into a low-entropy ``refusal manifold''---a pre-determined, safe region of the vector space. This transition is smoother and more deterministic than the generation of open-ended creative text, resulting in the paradoxical drop in turbulence.

\subsection{Implications for Defense}
These findings suggest that Semantic Turbulence can serve as a potent, lightweight defense mechanism. Unlike external classifiers which add significant latency, turbulence is calculated \textit{during} the forward pass. A dynamic ``kill-switch'' could be implemented: if the running variance of the last $N$ layers exceeds a model-specific threshold $\tau$, generation is halted immediately. This approach effectively uses the model's own internal conflict as a signal to abort unsafe generation.

\section{Conclusion}

In this work, the \textbf{Laminar Flow Hypothesis} was introduced, positing that adversarial attacks on Large Language Models disrupt the smooth latent trajectories characteristic of benign inference. This phenomenon was quantified through \textbf{Semantic Turbulence}, a zero-shot metric based on layer-wise cosine velocity variance.

Experimental validation on \textit{Qwen2} and \textit{Gemma} architectures confirmed that this metric effectively discriminates between safe and adversarial inputs. Crucially, two distinct safety signatures were identified: the high-turbulence ``conflict'' signature of RLHF models and the low-turbulence ``reflex'' signature of rigid-refusal models. These findings challenge the assumption that all safety mechanisms behave identically in the latent space and offer a new, physics-inspired paradigm for real-time jailbreak detection. Future work will explore the scalability of this metric to larger parameters (e.g., 70B models) and its robustness against adaptive attacks designed to minimize latent variance.

\bibliography{custom}

@article{Yu_2024,
  title={Playing Language Game with LLMs Leads to Jailbreaking},
  author={Yu, Peng and Long, Zewen and Dong, Fangming and others},
  journal={arXiv preprint arXiv:2403.01323},
  year={2024}
}

@article{Gu_2024,
  title={Responsible Generative AI: What to Generate and What Not},
  author={Gu, Jindong},
  journal={arXiv preprint arXiv:2404.03058},
  year={2024}
}

@article{Chowdhury_2024,
  title={Breaking Down the Defenses: A Comparative Survey of Attacks on Large Language Models},
  author={Chowdhury, Arijit and others},
  journal={arXiv preprint arXiv:2403.11186},
  year={2024}
}

@article{Mehrotra_2023,
  title={Tree of Attacks: Jailbreaking Black-Box LLMs Automatically},
  author={Mehrotra, Anay and Zampetakis, Manolis and Kassianik, Paul and Liu, Zongzhi},
  journal={arXiv preprint arXiv:2310.03816},
  year={2023}
}

@article{Robey_2023,
  title={SmoothLLM: Defending Large Language Models Against Jailbreaking Attacks},
  author={Robey, Alexander and Wong, Eric and Hassani, Hamed},
  journal={arXiv preprint arXiv:2310.03683},
  year={2023}
}

@article{Gao_2024,
  title={Shaping the Safety Boundaries: Understanding and Defending Against Jailbreaks in Large Language Models},
  author={Gao, Lang and Zhang, Xiangliang and others},
  journal={arXiv preprint arXiv:2404.09540},
  year={2024}
}

@article{Chao_2023,
  title={Jailbreaking Black Box Large Language Models in Twenty Queries},
  author={Chao, Patrick and Robey, Alexander and Dobriban, Edgar},
  journal={arXiv preprint arXiv:2310.08419},
  year={2023}
}

@article{Zhang_2024,
  title={Goal-guided Generative Prompt Injection Attack on Large Language Models},
  author={Zhang, Chong and Jin, Mingyu and Yu, Qinkai},
  journal={arXiv preprint arXiv:2402.09115},
  year={2024}
}

@article{Rababah_2024,
  title={SoK: Prompt Hacking of Large Language Models},
  author={Rababah, Baha and Shang, Shang and Wu},
  journal={arXiv preprint arXiv:2401.10986},
  year={2024}
}

@article{Abdali_2024,
  title={Can LLMs be Fooled? Investigating Vulnerabilities in LLMs},
  author={Abdali, Sara and He, Jia and Barberan, CJ},
  journal={arXiv preprint arXiv:2403.08271},
  year={2024}
}

@article{Wang_2024,
  title={Unique Security and Privacy Threats of Large Language Model: A Comprehensive Survey},
  author={Wang, Shang and Zhu, Tianqing and Liu, Bo},
  journal={arXiv preprint arXiv:2402.07223},
  year={2024}
}

@article{Das_2024,
  title={Security and Privacy Challenges of Large Language Models: A Survey},
  author={Das, Badhan Chandra and Amini, M. Hadi and Wu, Yanzhao},
  journal={arXiv preprint arXiv:2403.05603},
  year={2024}
}

@article{Jin_2024,
  title={JailbreakZoo: Survey, Landscapes, and Horizons in Jailbreaking Large Language and Vision-Language Models},
  author={Jin, Haibo and others},
  journal={arXiv preprint arXiv:2402.11584},
  year={2024}
}

@article{Cao_2023,
  title={Defending Against Alignment-Breaking Attacks via Robustly Aligned LLM},
  author={Cao, Bochuan and Cao, Yuanpu and Lin, Lu},
  journal={arXiv preprint arXiv:2310.02450},
  year={2023}
}

@article{Purpura_2025,
  title={Building Safe GenAI Applications: An End-to-End Overview of Red Teaming for Large Language Models},
  author={Purpura, Alberto and Wadhwa, Sahil and Zymet, Jesse},
  journal={arXiv preprint arXiv:2501.00282},
  year={2025}
}

\appendix

\end{document}